\documentclass[12pt,longbibliography,eprint]{revtex4-1}
\usepackage{epigraph}
\usepackage{graphicx}
\usepackage{epstopdf}
\usepackage{hyperref}
\usepackage{numprint}

\begin{document}
\preprint{V.M.}
\title{Multiple--Instance Learning: Radon--Nikodym Approach to Distribution Regression Problem.
}
\author{Vladislav Gennadievich \surname{Malyshkin}} 
\email{malyshki@ton.ioffe.ru}
\affiliation{Ioffe Institute, Politekhnicheskaya 26, St Petersburg, 194021, Russia}

\date{November, 27, 2015}

\begin{abstract}
\begin{verbatim}
$Id: DistReg1Step.tex,v 1.41 2015/12/02 11:00:50 mal Exp $
\end{verbatim}
For distribution regression problem,
where a bag of $x$--observations is mapped to a single $y$ value,
a one--step solution is proposed.
The problem of random distribution to
random value is transformed to
random vector to random value by taking
distribution moments of $x$ observations in a bag as random vector.
Then Radon--Nikodym or least squares theory
can be applied, what give $y(x)$ estimator.
The probability distribution of $y$ is also obtained,
what requires solving generalized eigenvalues problem,
matrix spectrum (not depending on $x$) give possible $y$ outcomes
and depending on $x$ probabilities of outcomes can be obtained by projecting
the distribution with fixed $x$ value (delta--function) to corresponding eigenvector.
A library providing numerically stable polynomial basis
for these calculations is available, what make the proposed approach practical.
\end{abstract}

\keywords{Distribution Regression, Radon--Nikodym}
\maketitle
\hfill\hbox{\small Dedicated to Ira Kudryashova}

\section{\label{intro}Introduction}
Multiple instance learning\cite{dietterich1997solving} is an important Machine Learning (ML)
concept having numerous applications\cite{yang2005review}.
In multiple instance learning class label is associated not with a
single observation, but with a ``bag'' of observations.
A very close problem is distribution regression
problem, where a sample distribution of $x$
is mapped to a single $y$ value.
There are numerous heuristics methods developed from both: ML and distribution regression sides, see \cite{zhou2004multi,szabo2014learning} for review.

As in any ML problem the most important part is not so much
the learning algorithm, but the way how the learned knowledge is represented.
Learned knowledge is often represented as a set of propositional rules,
regression function,
Neural Network weights, etc.
In this paper we consider the case where knowledge
is represented as a function of distribution moments.
Recent progress in numerical stability
of high order moments calculation\cite{2015arXiv151005510G}
allow the moments of very high order to be calculated,
e.g. in  Ref. \cite{2015arXiv151101887G} up to hundreds,
thus make this approach practical.

Most of distribution regression algorithms deploy a
two--step type of algorithm\cite{szabo2014learning}
to solve the problem.
In our previous work \cite{2015arXiv151107085G}
a two--step solution with knowledge representation in a form  of
Christoffel function was developed.
However, there is exist a one--step solution
to distribution regression problem, a random distribution to random value,
that converts each bag's observations to
moments of it, then solving the problem
random vector (the moments of random distribution) to random value.
Once this transition is made  an answer
of least squares or Radon--Nikodym
type from Ref. \cite{2015arXiv151005510G} can be applied
and close form result obtained.
The distribution of outcomes, if required, can be obtained
by solving generalized eigenvalues problem,
then matrix spectrum give possible $y$ outcomes,
and the square of projection
of localized at given $x$ bag distribution
to eigenvector give each outcome probability.
This matrix spectrum ideology is
similar to the one we used in  \cite{2015arXiv151107085G},
but is more generic and not reducible to Gauss quadrature.

The paper is organized as following:
In Section \ref{christoffel1Step}
a general theory of distribution regression is discussed and
close form result or least squares and Radon--Nikodym type are presented.
Then in Section \ref{christoffel1StepNum} an algorithm
is described and numerical example of
calculations is presented.
In Section \ref{christoffeldisc}
possible further  development is discussed.

\section{\label{christoffel1Step}One--Step Solution}

Consider distribution regression problem
where a bag of $N$ observations of $x$
is mapped to a single outcome observation $y$ for $l=[1..M]$.
\begin{eqnarray}
  (x_1,x_2,\dots,x_j,\dots,x_N)^{(l)}&\to&y^{(l)}  \label{regressionproblem}
\end{eqnarray}
A distribution regression problem can have a goal to estimate
$y$, average of $y$, distribution of $y$, etc.
given specific value of $x$

For further development we need $x$ basis $Q_k(x)$
and some $x$ and $y$ measure.
For simplicity, not reducing the generality of the approach,
we are going to assume that $x$ measure is a
sum over $j$ index  $\sum_{j=1}^{N}$,  $y$ measure is a $\sum_{l=1}^{M}$,
the basis functions $Q_k(x)$ are polynomials $k=0..d_x-1$,
where $d_x$ is the number of elements in $x$ basis, typical value for $d_x$ is below 10--15.

Let us convert the problem ``random distribution'' to ``random variable''
to the problem ``vector of random variables'' to ``random variable''.
The simplest way to obtain ``vector of random variables''
from $x_j^{(l)}$ distributions is to take the moments of it.
Now the $<Q_k>^{(l)}$ would be this random vector:
\begin{eqnarray}
  &&<Q_k>^{(l)}=\sum_{j=1}^{N} Q_k(x_j^{(l)}) \label{xmu} \\
 &&\left(<Q_0>^{(l)},\dots,<Q_{d_x-1}>^{(l)}\right) \to y^{(l)} \label{Qregressionproblem}
  \label{momy}
\end{eqnarray}
Then the (\ref{Qregressionproblem}) becomes vector to value problem.
Introduce
\begin{eqnarray}
  Y_q&=&\sum_{l=1}^{M} y^{(l)} <Q_q>^{(l)} \label{Yq} \\
  \left(G\right)_{qr}&=&\sum_{l=1}^{M} <Q_q>^{(l)} <Q_r>^{(l)} \label{Gramm} \\
  \left(yG\right)_{qr}&=&\sum_{l=1}^{M} y^{(l)} <Q_q>^{(l)} <Q_r>^{(l)} \label{GrammY}
\end{eqnarray}
The problem now is to estimate $y$ (or distribution of $y$)
given $x$ distribution, now mapped to a vector of moments $<Q_k>$ calculated on this
$x$ distribution.
Let us denote these input moments as  $M_k$
to avoid confusion with measures on $x$ and $y$.
For the case we study the $x$ value is given, and for a state with exact $x$  the $M_k$ values are:
\begin{eqnarray}
  M_k(x)&=& N Q_k(x) \label{Mdist}
\end{eqnarray}
what means that all $N$ observations in a bag give exactly the same $x$ value.
The problem now becomes a standard: random vector to random variable.
We have solutions of two types for this problem, see \cite{2015arXiv151005510G} Appendix D,
Least Squares $A_{LS}$ and Radon--Nikodym $A_{RN}$. The answers would be:
\begin{eqnarray}
  A_{LS}(x)&=& \sum\limits_{q,r=0}^{d_x-1} M_q(x) \left(G\right)^{-1}_{qr} Y_r \label{ALS} \\
  A_{RN}(x)&=& \frac{\sum\limits_{q,r,s,t=0}^{d_x-1} M_q(x) \left(G\right)^{-1}_{qr} \left(yG\right)_{rs}
    \left(G\right)^{-1}_{st}  M_t(x)}
  {\sum\limits_{q,r=0}^{d_x-1}M_q(x) \left(G\right)^{-1}_{qr} M_r(x)} \label{ARN}
\end{eqnarray}
The (\ref{ALS}) is least squares answer to $y$ estimation given $x$.
The (\ref{ARN}) is Radon--Nikodym answer to $y$ estimation given $x$.
These are the two $y$ estimators at given $x$ for distribution regression problem \ref{regressionproblem}.
These answers can be considered as an extension of least squares and Radon--Nikodym type
of interpolation from value to value problem
to random distribution to random variable problem. In case $N=1$ the
$A_{LS}$ and $A_{RN}$ are reduced exactly to value to value problem
considered in Ref. \cite{2015arXiv151005510G}. Note, that the $A_{LS}(x)$ answer not necessary
preserve $y$ sign, but  $A_{RN}(x)$ always preserve $y$ sign,
same as in  value to value problem.

If $y$ distribution at given $x$ need to be estimated this problem
can also be solved. With one--step approach of this paper we do not need $Q_m(y)$ basis used in two--step approach of  Ref. \cite{2015arXiv151107085G}
and outcomes of $y$ are estimated from $x$ moments only.
Generalized eigenvalues problem\cite{2015arXiv151005510G} give the answer:
\begin{eqnarray}
  \sum\limits_{r=0}^{d_x-1}\left(yG\right)_{qr} \psi^{(i)}_r &=& y^{(i)} \sum\limits_{r=0}^{d_x-1}\left(G\right)_{qr} \psi^{(i)}_r
  \label{gevproblem}
\end{eqnarray}
The result of (\ref{gevproblem}) is eigenvalues $y^{(i)}$ (possible outcomes)
and eigenvectors $\psi^{(i)}$ (can be used to compute the probabilities of outcomes).
The problem now becomes: given $x$ value estimate
possible $y$--outcomes and their probabilities.
The moments of states with given $x$ value are $NQ_q(x)$ from (\ref{Mdist}),
so the distribution with (\ref{Mdist}) moments should be projected
to distributions corresponding to $\psi^{(i)}_q$ states, the square of this projection
give the weight and normalized weight give the probability.
This is actually very  similar to ideology 
we used in \cite{2015arXiv151107085G}, but the eigenvalues from (\ref{gevproblem})
no longer have a meaning of Gauss quadrature nodes.
The eigenvectors $\psi^{(i)}_r$ correspond to distribution
with moments $<Q_q>=\sum_{r=0}^{d_x-1} \left(G\right)_{qr}  \psi^{(i)}_r$,
and the distribution with such moments correspond to $y^{(i)}$ value.
These distributions can be considered as
``natural distribution basis''. This is an important
generalizatioh
of Refs. \cite{2015arXiv151005510G,2015arXiv151101887G} approach to random distribution,
where
natural basis for random value, not random distribution,
was considered.

The projection of two $x$ distributions with moments $M^{(1)}_k$ and $M^{(2)}_k$
on each other is
\begin{eqnarray}
  <M^{(1)}|M^{(2)}>_{\pi}&=&\sum_{q,r=0}^{d_x-1} M^{(1)}_q \left(G\right)^{-1}_{qr} M^{(2)}_r
  \label{proj}
\end{eqnarray}
then the required probabilities, calculated by projecting the (\ref{Mdist})
 distribution to natural basis states,
are:
\begin{eqnarray}
  w^{(i)}(x)&=&\left(\sum_{r=0}^{d_x-1}M_r(x)  \psi^{(i)}_r \right)^2
  \label{wi}
  \\
  P^{(i)}(x)&=&w^{(i)}(x)/\sum_{r=0}^{d_x-1} w^{(r)}(x) \label{Pi}
\end{eqnarray}
The (\ref{gevproblem}) and (\ref{Pi}) is  one--step answer to distribution regression problem:
find the outcomes $y^{(i)}$ and their probabilities $P^{(i)}(x)$.
Note, that in this setup possible outcomes $y^{(i)}$ do not depend on $x$,
and only probabilities $P^{(i)}(x)$ of outcomes  depend on $x$.
This is different from a two--step solution of \cite{2015arXiv151107085G}
where outcomes and their probabilities both depend on $x$.
Also note that
$\sum_{r=0}^{d_x-1} w^{(r)}(x)=\sum_{q,r=0}^{d_x-1} M_q(x) \left(G\right)^{-1}_{qr} M_r(x)$.

One of the major difference between the probabilities (\ref{Pi})
and probabilities from Christoffel function approach \cite{2015arXiv151107085G}
is that the (\ref{Pi}) has a meaning of ``true'' probability
while in two--step solution \cite{2015arXiv151107085G} Christoffel function value is used as a proxy to probability on first step.
It is important to note how the knowledge is represented in these models.
The model (\ref{ALS}) has learned knowledge represented in $d_x$ by $d_x$ matrix
(\ref{Gramm}) and $d_x$ size vector (\ref{Yq}).
The model (\ref{ARN}) as well as distribution answer (\ref{Pi})
has learned knowledge represented in two $d_x$ by $d_x$ matrices
(\ref{Gramm}) and (\ref{GrammY}).

\section{\label{christoffel1StepNum}Numerical estimation of One--Step Solution}
Numerical instability
similar to the one of two--stage Christoffel function approach \cite{2015arXiv151107085G}
also arise for approach in study, but now the situation is much
less problematic, because
we do not have $y$--basis $Q_m(y)$, and all the dependence
on $y$ enter the answer through matrix (\ref{GrammY}).
In this case the only stable  $x$
basis $Q_k(x)$ is required.

The algorithm for  $y$ estimators of (\ref{ALS}) or (\ref{ARN})  is this:
Calculate $<Q_k>^{(l)}$ moments from (\ref{xmu},
then calculate matrices (\ref{Gramm}) and (\ref{GrammY}),
if least squares approximation is required also calculate moments (\ref{Yq}).
In contrast with Christoffel function approach
where $<Q_qQ_r>; q,r=[0..d_x-1]$ matrix
can be obtained from $Q_k ; k=[0..2d_x-1]$ moments
by application of polynomials multiplication operator,
here the (\ref{Gramm}) and (\ref{GrammY}) can be hardly obtained this
way for $N>1$ and should be calculated directly from sample. This
is not a big issue, because $d_x$ is typically not large.
Then inverse matrix $\left(G\right)_{qr}$ from (\ref{Gramm}),
this matrix is some kind similar to Gramm matrix, but uses
distribution moments, not basis functions.
Finally put all these to (\ref{ALS}) for least squares $y(x)$ estimation
or to (\ref{ARN}) for Radon--Nikodym $y(x)$ estimation.

If $y$-- distribution is required then solve
generalized eigenvalues problem (\ref{gevproblem}),
obtain $y^{(i)}$ as possible $y$--outcomes (they do not depend on $x$),
and calculate $x$--dependent probabilities (\ref{Pi}), these are
squared projection coefficient of a state with specific $x$ value,
point--distribution (\ref{Mdist}),
or some other $x$ distribution of general form,
to $\psi^{(i)}$ eigenvector.

To show an application of this approach let us take several simple
distribution to apply the theory. Let $\epsilon$ be a uniformly distributed $[-1;1]$
random variable and take $N=1000$ and $M=10000$.
Then consider sample distributions build as following
1) For $l=[1..M]$ take random $x$ out of $[-1;1]$ interval.
2) Calculate $y=f(x)$, take this $y$ as $y^{(l)}$.
3) Build a bag of $x$ observations as $x_j=x+R\epsilon ; j=[1..N]$, where $R$ is a parameter.
The following three $f(x)$ functions for building sample distribution are used:
\begin{eqnarray}
  f(x)&=&x \label{flin} \\
  f(x)&=&\frac{1}{1+25x^2} \label{frunge} \\
  f(x)&=&\left\{\begin{array}{ll} 0 & x\le 0 \\ 1 & x>0\end{array}\right.
  \label{fstep}
\end{eqnarray}

\begin{figure}[t]
\includegraphics[width=8cm]{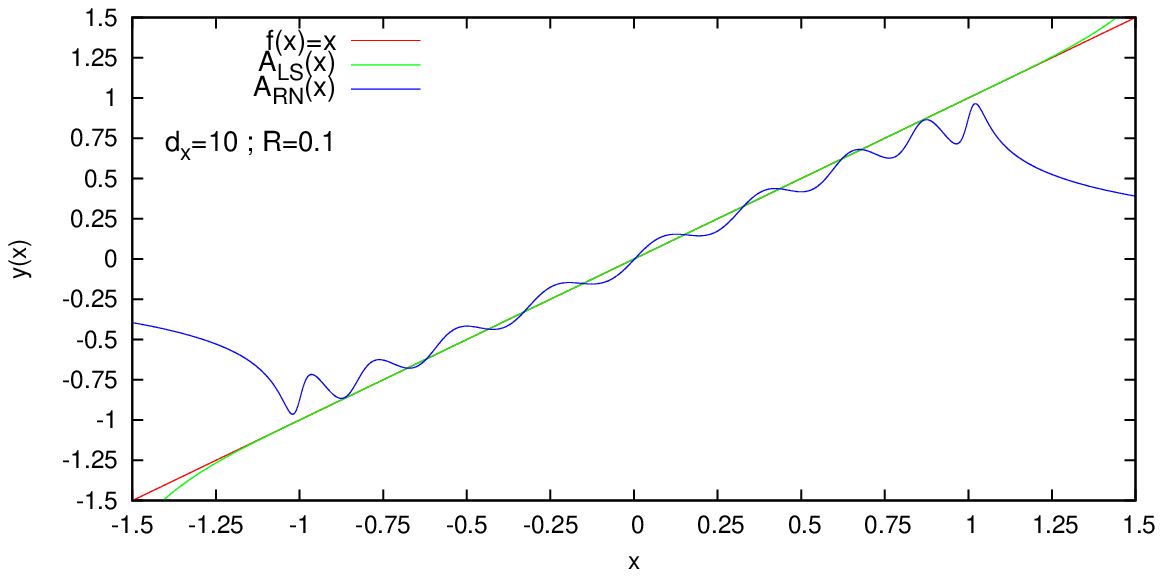}
\includegraphics[width=8cm]{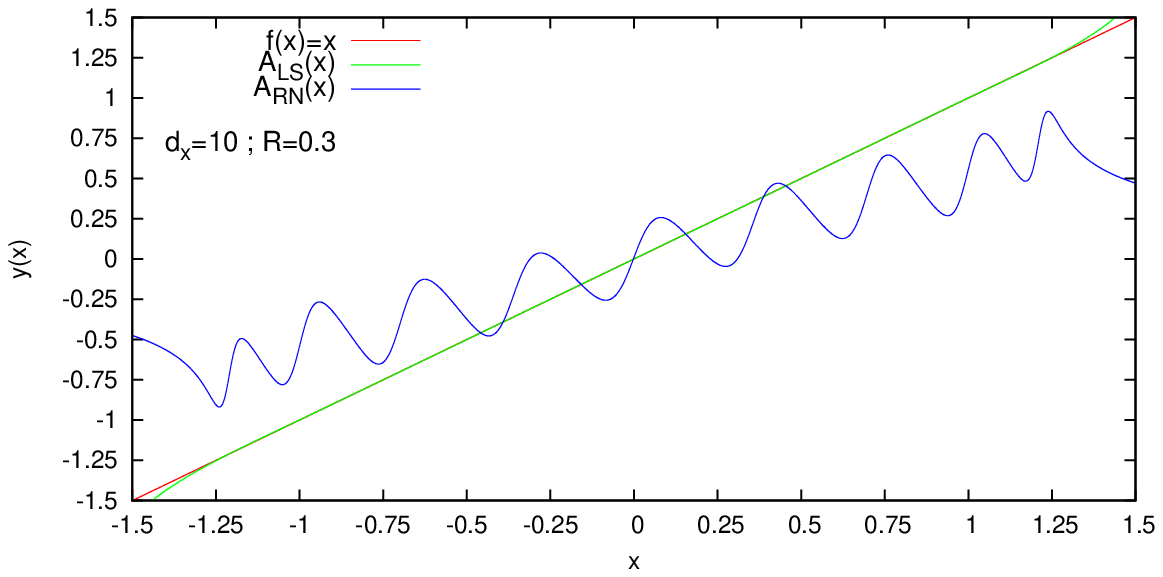}
\includegraphics[width=8cm]{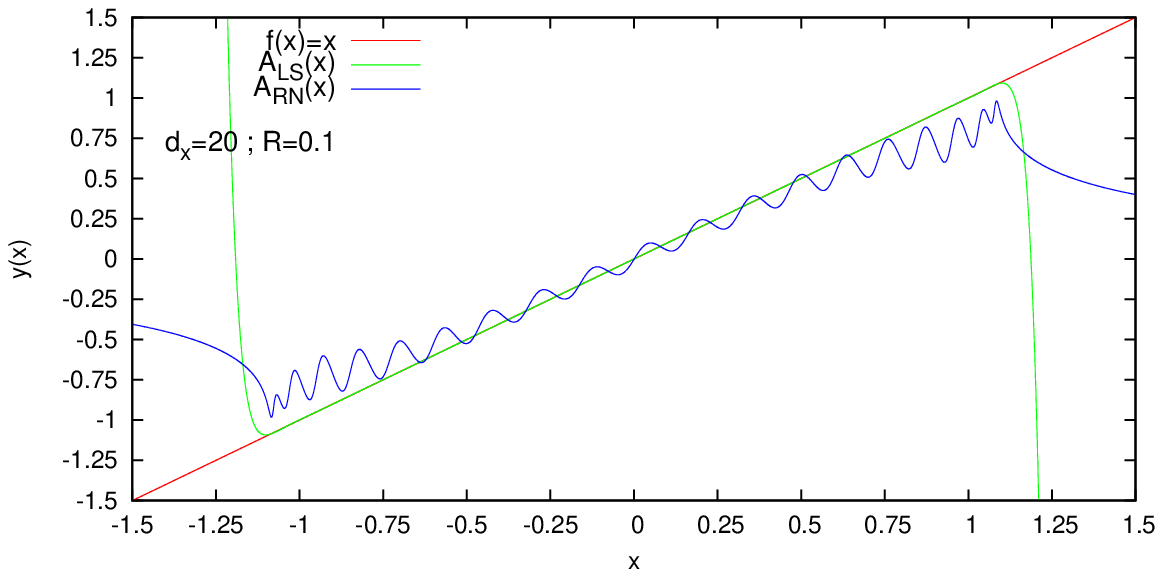}
\includegraphics[width=8cm]{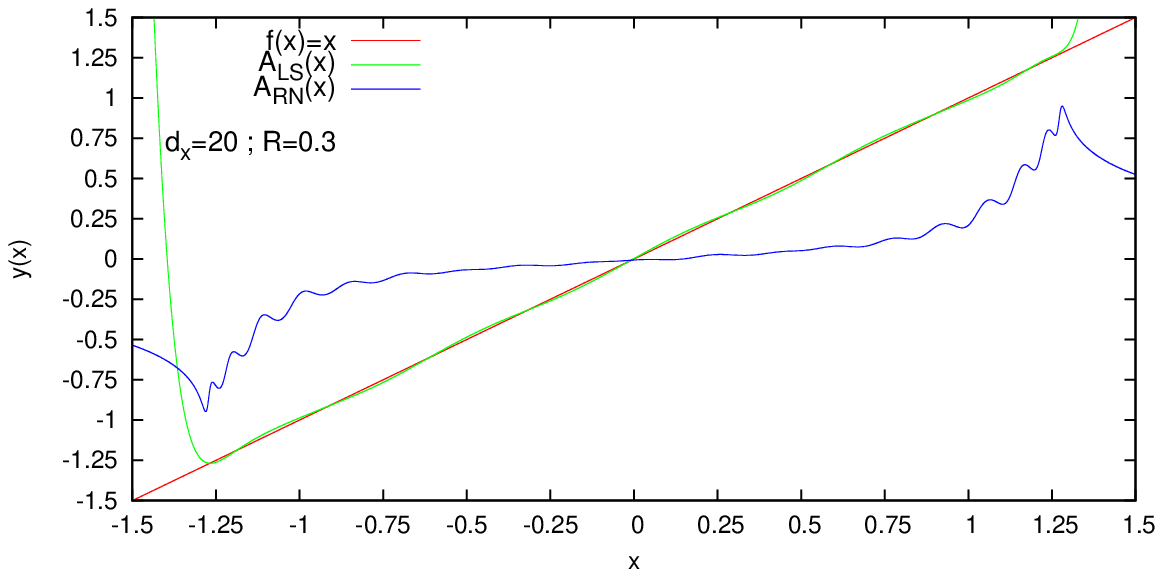}
\caption{\label{fig:flin}
  The $y(x)$ estimation for $f(x)$ from (\ref{flin}).
  }
\end{figure}

\begin{figure}[t]
\includegraphics[width=8cm]{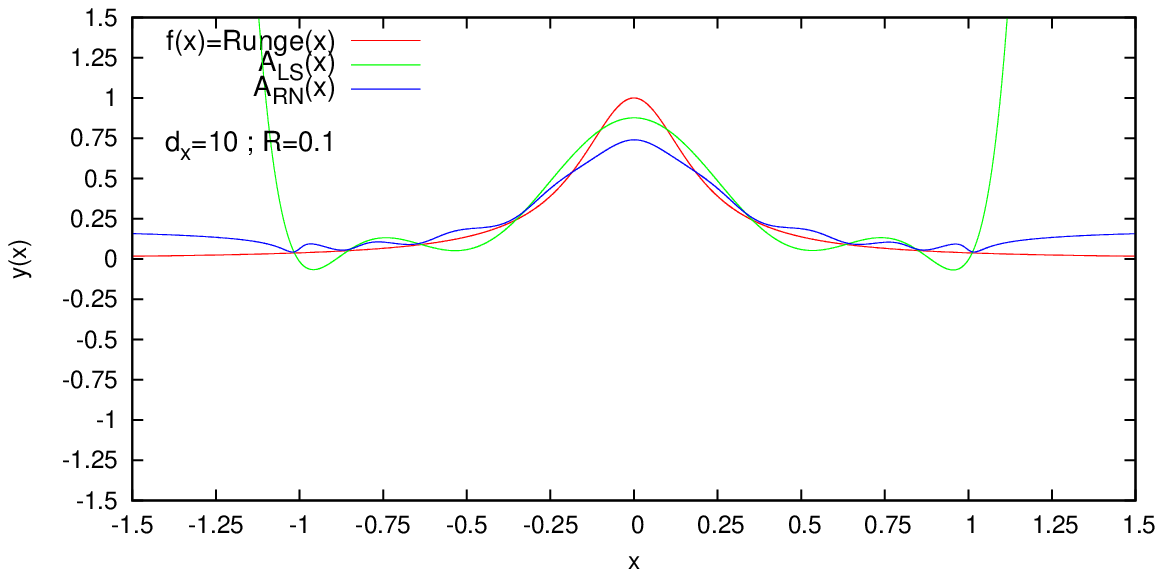}
\includegraphics[width=8cm]{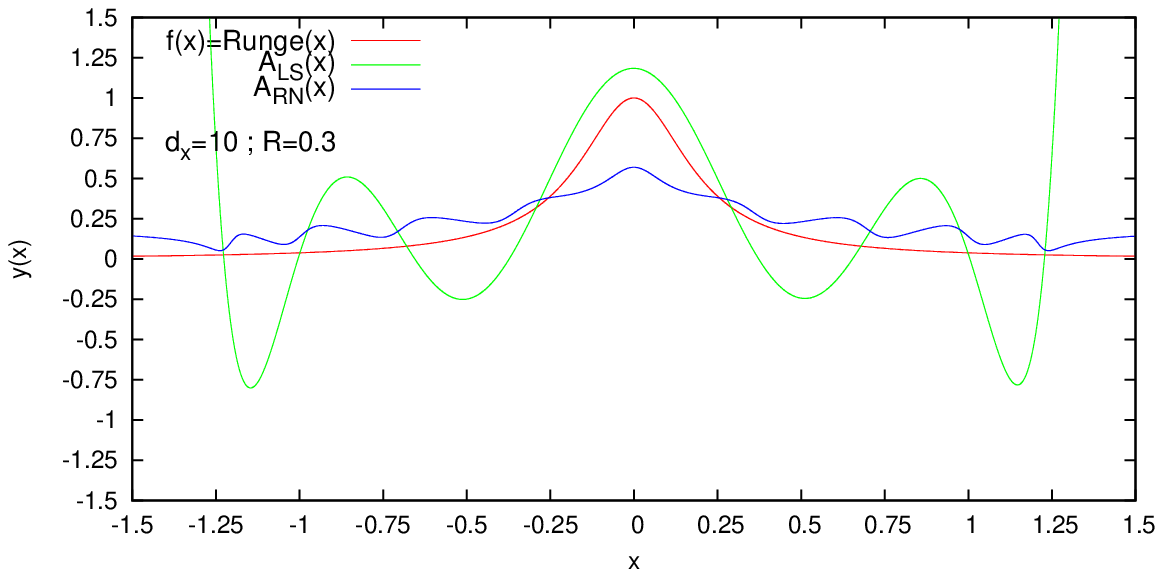}
\includegraphics[width=8cm]{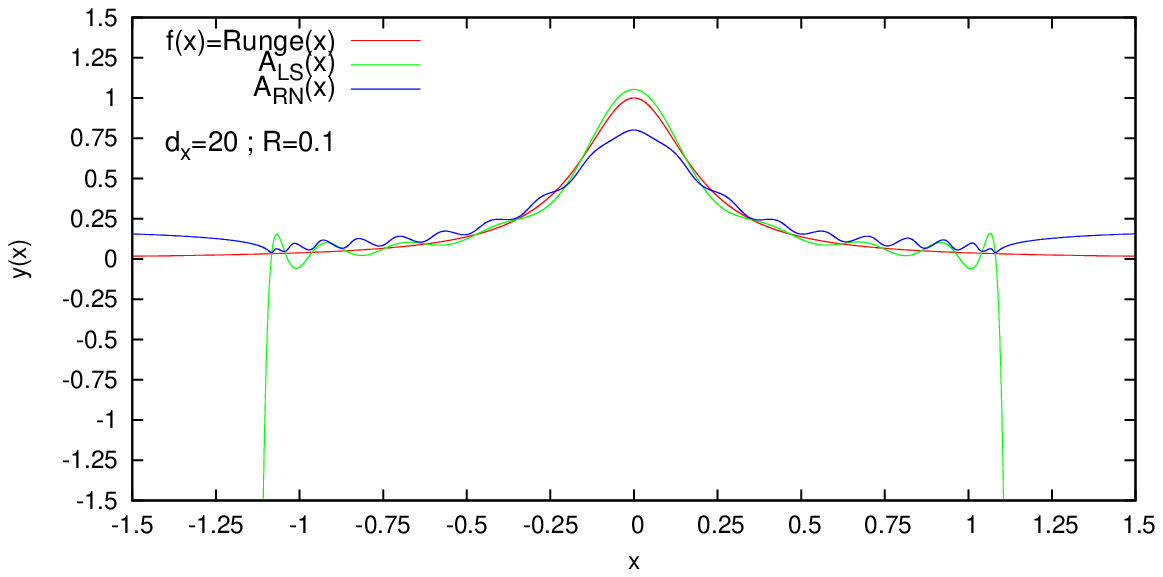}
\includegraphics[width=8cm]{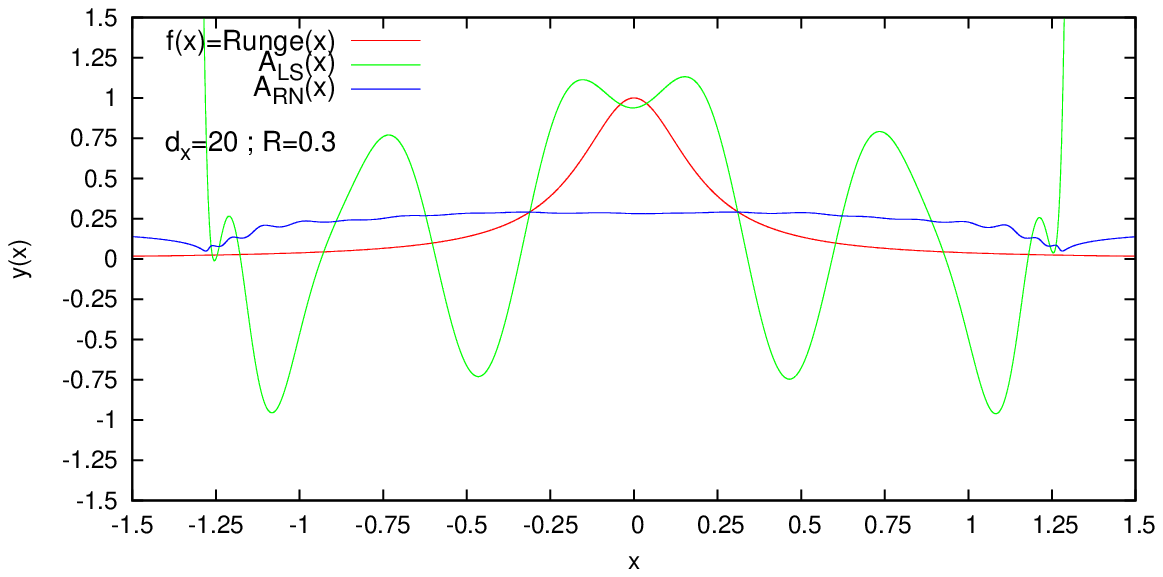}
\caption{\label{fig:frunge}
  The $y(x)$ estimation for $f(x)$ from (\ref{frunge}).
  }
\end{figure}

\begin{figure}[t]
\includegraphics[width=8cm]{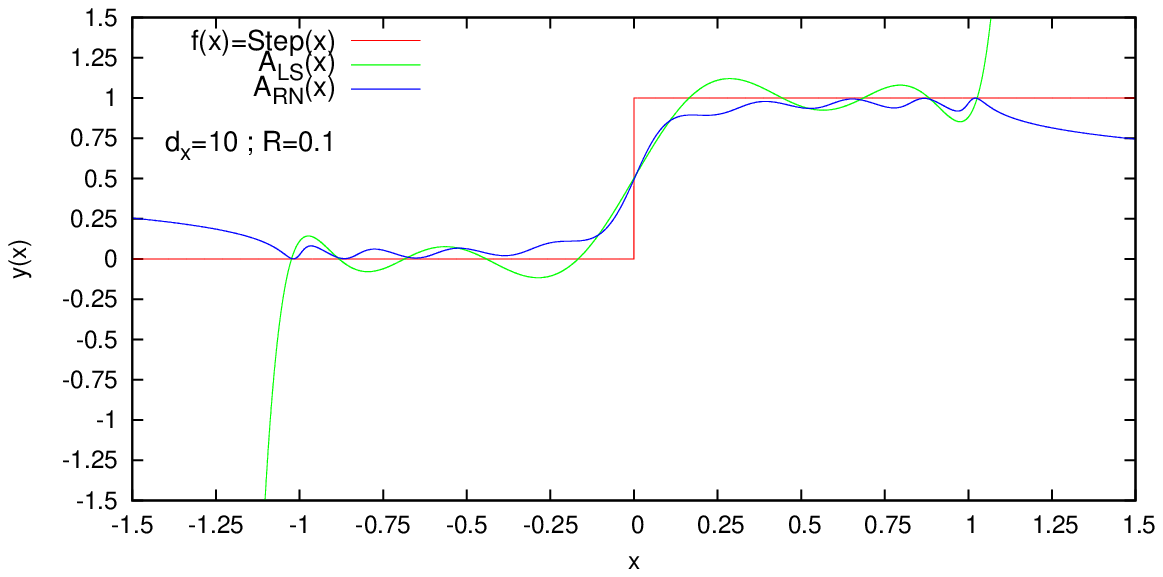}
\includegraphics[width=8cm]{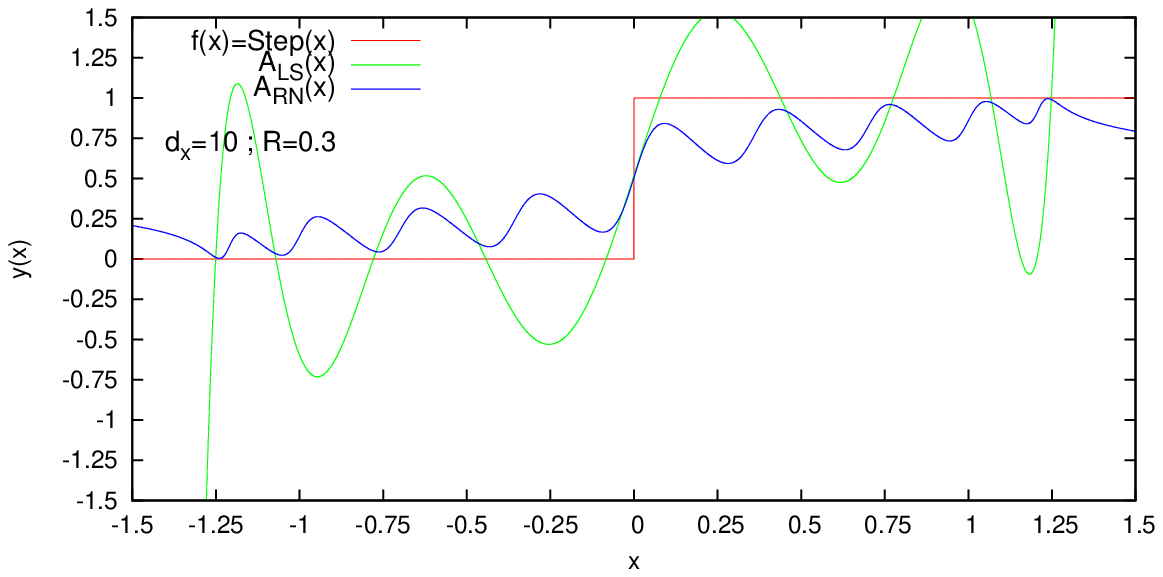}
\includegraphics[width=8cm]{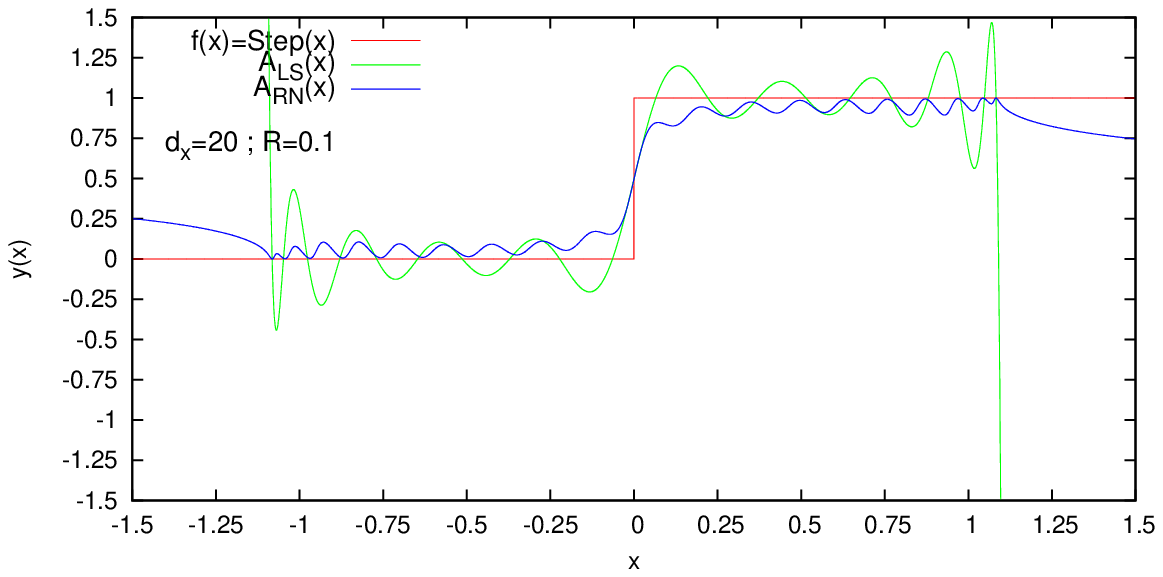}
\includegraphics[width=8cm]{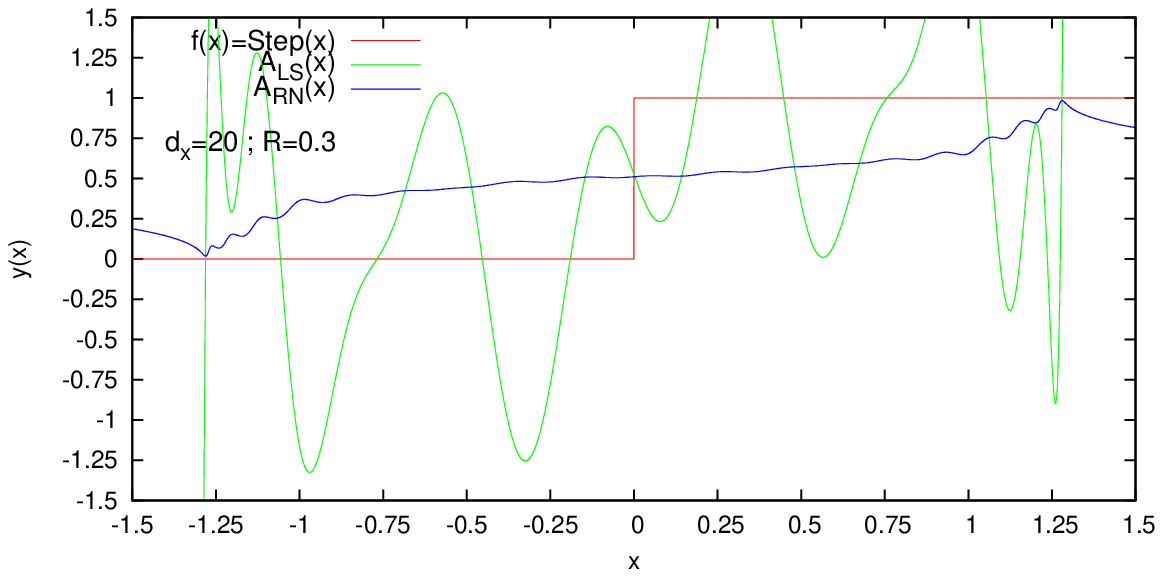}
\caption{\label{fig:fstep}
  The $y(x)$ estimation for $f(x)$ from (\ref{fstep}).
  }
\end{figure}

In Figs. \ref{fig:flin},  \ref{fig:frunge}, \ref{fig:fstep},
the  (\ref{ALS}) and (\ref{ARN}) answers
are presented for $f(x)$ from (\ref{flin}), (\ref{frunge}) and (\ref{fstep})
respectively 
for $R=\{0.1,0.3\}$
and $d_x=\{10,20\}$. The $x$ range is specially taken slightly
wider that $[-1; 1]$ interval to see possible divergence
outside of measure support.
In most cases Radon--Nikodym answer is superior,
and in addition to that it preserves the sign of $y$.
Least squares approximation is good for special case $f(x)=x$
and typically diverges at $x$ outside of measure support.

\begin{figure}[t]
\includegraphics[width=14cm]{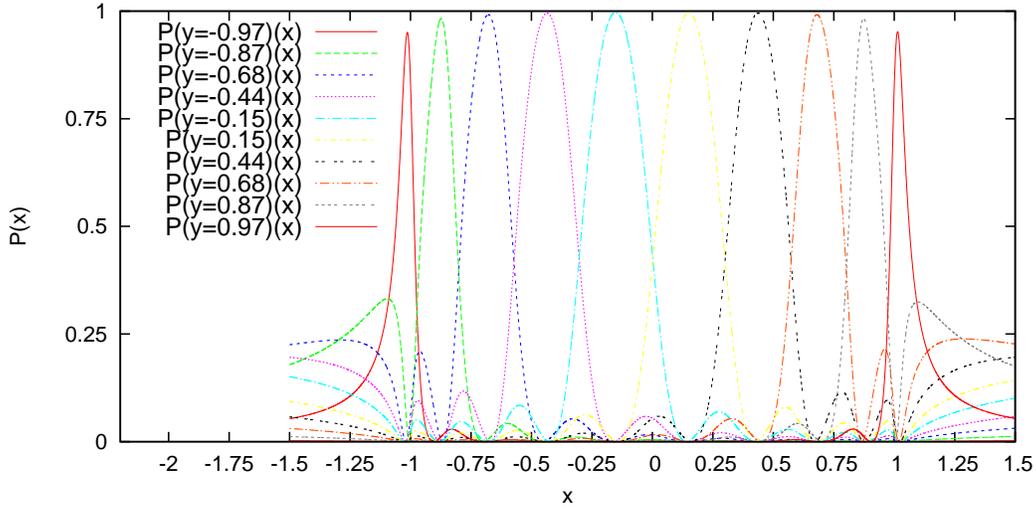}
\caption{\label{fig:Px}
  Probabilities for all $d_x=10$ outcomes of $y^{(i)}$ as a function of $x$ for $f(x)$ from (\ref{flin}).
  }
\end{figure}
The numerical estimation of probability function
(the $y^{(i)}$ and $P^{(i)}(x)$) were also calculated
and eigenvalue index $i$, corresponding to maximal $P$
typically correspond to $y^{(i)}$, for which $f(x)$ is most close.
For simplistic case (\ref{flin}) see Fig. \ref{fig:Px}.
See the Ref. \cite{polynomialcode}, file
com/polytechnik/ algorithms/ ExampleDistribution1Stage.scala
for algorithm implementation.

\section{\label{christoffeldisc}Discussion}
In this work a one--stage approach is applied to
distribution regression problem. The bag's observations
are initially converted to moments, then
least squares or Radon--Nikodym theory can be applied and closed
form answer to be received.
These (\ref{ALS}) and (\ref{ARN}) estimate $y$ value given $x$.
This answer can be generalized to
``what is $y$ estimate given distribution of $x$''.
For this problem obtain moments $<Q_k>$, corresponding to given distribution of $x$, first, then use them in
(\ref{ALS}) or (\ref{ARN}) instead of $M_k(x)$, corresponding
to localized at $x$ state. Similary,
if  probabilities of $y$ outcomes are required
for given distribution of $x$, the  $<Q_k>$
should be used in weights expression (\ref{wi})
instead of $M_k(x)$ (this is a special case of two
distribution projection on each other
(\ref{proj})).
Computer code implementing the algorithms is available\cite{polynomialcode}.

And in conclusion we want to discuss possible directions of future development.
\begin{itemize}
  \item
In this work a closed form solution 
for random distribution to random value problem (\ref{regressionproblem})
is found. The question arise about problem order increase,
replace ``random distribution'' by
``random distribution of random distribution'' (or even further
``random distribution of random distribution of random distribution'', etc.).
In this case each $x_j$ in (\ref{regressionproblem})
should be treated as a  sample distribution itself,
and index $j$ then can be treated as 2D index $x_{j_1,j_2}$.
Working with 2D indexes is actually very similar
to working with images, see Ref. \cite{2015arXiv151101887G}
where the 2D index was used for image reconstruction
by applying Radon--Nikodym or least squares approximation.
Similarly, the results of this paper,
can be generalized to higher order problems,
by considering all indexes as 2D.

\item Obtaining possible $y$ outcomes as matrix spectrum  (\ref{gevproblem})
  and then calculating their probabilities by projection (\ref{proj}) of given distribution (point distribution (\ref{Mdist}) is a simplest example of such)
  to eigenvectors (\ref{Pi}) is a powerful
  approach to estimation of $y$ distribution under given condition.
  We can expect this approach to show
  good performance for data drawn from
  a wide range of probability distributions,
  especially for distributions that are not normal. The reason
  is because the (\ref{gevproblem}) is expressed in terms of probability states,
  what make the role of outliers much less important, compared to
  methods based on $L^2$ norm, particulary least squares.
  For example, this approach can be applied to distribtions
  where only first moment of $y$ is finite, while the $L^2$ norm
  approaches require second moment of $y$ to be finite,
  what make them inapplicable to distributions with infinite
  standard deviation.
  We expect the (\ref{gevproblem}) approach can be
  a good foundation for construction of Robust Statistics\cite{huber2011robust}.
\end{itemize}

\bibliography{LD}

\end{document}